\documentclass[10pt]{article} % For LaTeX2e
\usepackage[preprint]{tmlr}
\usepackage{amsmath,amsfonts,bm}

\def\eqref#1{equation~\ref{#1}}
\def\1{\bm{1}}

\DeclareMathAlphabet{\mathsfit}{\encodingdefault}{\sfdefault}{m}{sl}
\SetMathAlphabet{\mathsfit}{bold}{\encodingdefault}{\sfdefault}{bx}{n}

\usepackage{hyperref}
\usepackage{url}
\usepackage{graphicx} %ADDED !
\usepackage{booktabs} % ADDED !
\usepackage{multicol} % ADDED !
\usepackage{multirow} % ADDED !
\usepackage{amsthm}  % ADDED !
\usepackage{amsmath,amssymb}

\usepackage{soul}
\usepackage[dvipsnames]{xcolor}

\title{Jump Start or False Start? A Theoretical and Empirical Evaluation of LLM-initialized Bandits}

\author{\name Adam Bayley \email 19ahb@queensu.ca \\
      \addr Department of Electrical and Computer Engineering\\
      Queen's University 
      \AND
      \name Xiaodan Zhu \email xiaodan.zhu@queensu.ca \\
      \addr Department of Electrical and Computer Engineering \\
      Queen's University
      \AND
      \name Raquel Aoki \email raquel.aoki@borealisai.com \\
      \addr RBC Borealis
      \AND 
      \name Yanshuai Cao \email yanshuai.cao@borealisai.com \\
      \addr RBC Borealis
      \AND
      \name  Kevin H. Wilson \email kevin.h.wilson@borealisai.com \\
      \addr RBC Borealis
}

\newtheorem{thm}{Theorem}

\def\month{MM}  % Insert correct month for camera-ready version
\def\year{2025} % Insert correct year for camera-ready version
\def\openreview{\url{https://openreview.net/forum?id=XXXX}} % Insert correct link to OpenReview for camera-ready version

\begin{document}

\maketitle

\begin{abstract}

The recent advancement of Large Language Models (LLMs) offers new opportunities to generate user preference data to warm-start bandits. Recent studies on contextual bandits with LLM initialization (CBLI) have shown that these synthetic priors can significantly lower early regret. However, these findings assume that LLM-generated choices are reasonably aligned with actual user preferences. In this paper, we systematically examine how LLM-generated preferences perform when random and label-flipping noise is injected into the synthetic training data. For aligned domains, we find that warm-starting remains effective up to 30\% corruption, loses its advantage around 40\%, and degrades performance beyond 50\%. When there is systematic misalignment, even without added noise, LLM-generated priors can lead to higher regret than a cold-start bandit. To explain these behaviors, we develop a theoretical analysis that decomposes the effect of random label noise and systematic misalignment on the prior error driving the bandit’s regret, and derive a sufficient condition under which LLM-based warm starts are provably better than a cold-start bandit. We validate these results across multiple conjoint datasets and LLMs, showing that estimated alignment reliably tracks when warm-starting improves or degrades recommendation quality. 

%for intro, later: Our experiments can be run here: [LINK]. 

\end{abstract}
  
\section{Introduction}
\label{sec:Introduction}

As a novice attempts to solve a new problem without prior heuristics, the search for a solution often begins as a random walk. This lack of structural guidance mirrors the fundamental challenge in online learning, known as the ``cold start'' problem. When an agent is initialized \textit{tabula rasa}, without any preconceived notions, the agent faces a vast action space with no means to distinguish between optimal and suboptimal decisions. Consequently, the agent is forced into exploration, often incurring high sample complexity and significant performance penalties before converging to a competitive strategy. 

Contextual multi-armed bandits (CBs) have emerged as an essential tool to address this problem in the online learning setting. Here, an agent is tasked with choosing a piece of content for each user in a sequence based on the content's and users' features. The agent receives feedback (e.g., a click) usually instantaneously after choosing the content, and may use this feedback to update itself before choosing the next piece of content. By simultaneously balancing \textit{exploration} (gathering information about user preferences) and \textit{exploitation} (utilizing the information gathered to maximize some reward function), CBs optimize real-time recommendations \citep{li2010contextual} and admit a sublinear finite‑time regret bound under linear payoff assumptions \citep{chu2011contextual}. However, when CBs have yet to gather any user data, they perform essentially randomly and thus exhibit linear regret \citep{li2010contextual, aur2002analysisofmabproblem}. Traditional approaches have sought to address this limitation by warm-starting bandits using historical user data or expert knowledge \citep{pmlr-v97-zhang19b}.

The recent advancement of Large Language Models (LLMs) offers new opportunities and alternatives,  providing built-in knowledge about human preferences \citep{brown2020languagemodelsfewshotlearners}. \citet{alamdari2024jumpstartingbanditsllmgenerated} 
introduced the Contextual Bandits with LLM Initialization (CBLI) framework, which prompts an LLM to simulate user preferences to generate a synthetic pre-training dataset for a contextual bandit. By simulating bandit performance using data from a conjoint survey experiment, the authors showed that ``jump-starting'' a CB with this synthetic data achieved an impressive 14–20\% reduction in early regret. This demonstrated that even if the LLM-generated preferences are not perfectly accurate, they can still provide a much better starting point than no prior data. The key limitation of their work is that it does not address the underlying implicit assumptions, in addition to focusing only on a single domain. 

%\textbf{Motivation}. 
Previous studies have shown conflicting evidence on whether LLMs can accurately simulate human decision making \citep{10.1145/3442188.3445922,Kosinski_2024}. The original CBLI results implicitly rely on an alignment assumption: that LLM-simulated preferences are reasonably close to human preferences on the target task. Despite the demonstrated benefits of the CBLI framework, its robustness to bias and misalignment in these LLM-generated priors is insufficiently understood. Understanding when the framework may break down is critical before deployment in real-world systems.

In this paper, we present a theoretical and empirical study on LLM-generated priors for bandit algorithms. Consistent with the experimental protocol established in the original CBLI framework \citep{alamdari2024jumpstartingbanditsllmgenerated}, and the broader literature addressing the cold-start problem in personalized recommendation \citep{li2010contextual, zhou2016latentcontextualbanditsapplication}, we situate our work within the domain of recommender systems --- a core component of modern digital platforms designed to help users navigate environments characterized by extreme information overload. We specifically focus on contextual bandits (and their sleeping counterparts \citep{kanade2009sleeping}), as they provide a principled framework to integrate dynamic factors (e.g., time, location) that have been shown to significantly enhance recommendation quality \citep{panniello2009experimental}.

In summary, our contributions are threefold:

\begin{itemize}
    \item \textbf{Noisy-CBLI framework}: We introduce a novel extension to CBLI where synthetic noise is injected into the LLM-generated preference data before pretraining the bandit. We consider two noise injection strategies: (a) Random Replacement---replacing a certain percentage of LLM-generated responses with random choices, and (b) Preference Flipping---flipping the chosen option in a binary choice for a certain percentage of the responses. This framework allows us to simulate varying levels and types of LLM errors and study their impact.

    \item \textbf{Systematic noise impact evaluation}: We conduct an empirical study across three conjoint datasets and multiple LLMs to measure how noise and misalignment affect cumulative regret. For aligned domains, we find that warm-start gains persist up to roughly 30\% preference-flipping, vanish around 40\%, and reverse beyond 50\%, where synthetic priors become harmful, while bandits remain comparatively resilient to random replacement noise. In contrast, for tasks where LLM preferences are systematically misaligned with human responses, we show that CBLI can underperform a cold-start bandit even with no injected noise, highlighting alignment as the key failure mode.

    \item \textbf{Alignment-based theoretical analysis}: We develop a theoretical analysis of CBLI in a sleeping linear contextual bandit model with an LLM-induced prior. The analysis identifies a single prior-error term that captures the combined effect of random label noise and systematic misalignment between LLM-simulated and human rewards, and yields a sufficient condition under which LLM-based warm starts are likely to improve over a cold-start bandit. We show that this condition closely tracks the observed transition between regimes where CBLI helps and where it harms, thereby grounding the noisy-CBLI and misalignment results in a unified theoretical perspective.
  
\end{itemize}
  
\section{Related Works}
\label{sec:Related Works}

\textbf{Contextual and cold-start bandits}. Contextual and non-contextual bandits formalize online personalization under partial feedback, with linear methods such as LinUCB and related regret analyses forming a standard baseline \citep{li2010contextual, chu2011contextual,aur2002analysisofmabproblem}. As discussed above and following previous works and experimentation \citep{alamdari2024jumpstartingbanditsllmgenerated,zhou2016latentcontextualbanditsapplication,panniello2009experimental}, we situate our work in the domain of recommender systems. Sequence-aware and session-based models capture evolving user preferences over interaction histories but do not fully resolve user and item cold-start issues in deployed systems \citep{hidasi2016sessionbasedrecommendationsrecurrentneural,quadrana2018sequenceawarerecommendersystems}. Variants of contextual bandits explicitly targeting cold-start recommendation include latent contextual bandits for new-user personalization \citep{zhou2016latentcontextualbanditsapplication}, and broader overviews of such extensions are given in contextual bandit surveys such as \citet{zhou2016surveycontextualmultiarmedbandits}. Relatedly, settings with stochastic action availability motivate “sleeping” bandit formulations \citep{kanade2009sleeping}.

\textbf{Warm-starting and transfer in bandits}. A common approach to the cold-start problem is to initialize bandit learning with supervised or logged feedback. Robust warm-starting methods analyze regimes where offline and online signals diverge and propose procedures that mitigate harmful initialization \citep{pmlr-v97-zhang19b}. Transfer learning for contextual bandits similarly characterizes how source–target similarity governs whether transfer reduces regret or induces negative transfer \citep{cai2024transferlearningcontextualmultiarmed}. Related multi-task formulations treat transfer learning as shared structure across bandit tasks via hierarchical priors \citep{hong2022hierarchicalbayesianbandits}, while recent work explicitly targets negative transfer under covariate shift in latent and other contextual bandits \citep{deng2025transferlearninglatentcontextual}. 

\textbf{LLMs in recommendation and sequential decision-making}. More recently, LLMs have been used to improve recommenders by reframing recommendation as language modeling/prompting \citep{geng2023recommendationlanguageprocessingrlp,petrov2023generativesequentialrecommendationgptrec}, and by generating sequential recommendations autoregressively \citep{volodkevich2024autoregressivegenerationstrategiestopk}. LLMs have also been incorporated into contextual bandit pipelines as auxiliary signal providers \citep{baheri2023llmsaugmentedcontextualbandit}, and more broadly positioned as agents for sequential decision-making \citep{NEURIPS2023_f6b22ac3}. Conversational recommendation and LLM-centered systems have also shown promise \citep{gao2023chatrecinteractiveexplainablellmsaugmented,Bao_2023}, with instruction tuning and alignment methods providing mechanisms by which model outputs may approximate human feedback \citep{ouyang2022traininglanguagemodelsfollow, bai2022constitutionalaiharmlessnessai} supported by evidence on scaling and capability trends \citep{brown2020languagemodelsfewshotlearners, openai2024gpt4technicalreport}. 

\textbf{LLMs as preference simulators and synthetic label generators.} Beyond their role as representation learners and controllers, LLMs are increasingly used as simulated participants (``silicon samples'') in behavioral, social-science, and preference-elicitation studies \citep{Argyle_Busby_Fulda_Gubler_Rytting_Wingate_2023,https://doi.org/10.1002/mar.21982}. Evidence is mixed: on the one hand, LLMs can match some aggregate patterns in human judgments in specific settings, including theory-of-mind style tasks and certain interactive behaviors \citep{Kosinski_2024}, and have shown promise in jump-starting bandit recommenders with synthetic preference data \citep{alamdari2024jumpstartingbanditsllmgenerated}. On the other hand, multiple evaluations show systematic deviations that undermine naive ``drop-in'' use, including ordering/labeling artifacts and a tendency toward near-uniform or otherwise distorted response distributions once such artifacts are controlled \citep{dominguezolmedo2024questioningsurveyresponseslarge, 10.1145/3708319.3733685}. These concerns are amplified by the ``analytic flexibility'' of silicon-sample pipelines: small choices in prompting, sampling, or scoring can substantially change whether a model appears aligned with human data, with no single configuration performing well across evaluation criteria \citep{cummins2025threatanalyticflexibilityusing}. Broader critiques emphasize that scale and fluent outputs do not guarantee representativeness, transparency, or safety, motivating caution when substituting synthetic respondents for humans \citep{10.1145/3442188.3445922}.

\textbf{Positioning of Our Work}. Prior results demonstrate that prompting LLMs for synthetic conjoint choices can meaningfully reduce early regret when used as a warm-start prior, but they largely rely on an implicit alignment assumption between LLM-simulated and human preferences. At the same time, the broader ``LLMs as silicon samples'' literature reports mixed fidelity and substantial sensitivity to design choices, raising the concern that LLM-generated preference data may contain both unstructured mistakes and systematic deviations from the target population. We develop the noisy-CBLI framework to evaluate the robustness of LLM-generated priors for warm-starting under two types of corruption: random replacement (synonymous with uninformative feedback) and preference flipping (an example of a biased model). We separately formalize systematic misalignment through target shift and empirically delineate regimes in which warm-starting improves performance. Theoretically, we consolidate these effects through a prior-error term characterizing when warm-starting can outperform a cold-start, and we develop an alignment-based diagnostic that anticipates the transition from beneficial to harmful initialization.

\section{Methodology}
\label{methodology}

In this section, we build on the CBLI framework to study its robustness under noisy and potentially misaligned synthetic priors. We first describe three real‐world conjoint datasets used in our study, then formalize the contextual‐bandit problem and recap the CBLI jump-start method. Finally, we introduce two noise-injection strategies---random response replacement and preference flipping---that systematically corrupt the synthetic priors, defining the noisy CBLI variants we evaluate under realistic noisy conditions.

\subsection{Datasets}
\label{Datasets}

We use data collected from three conjoint surveys.  In each, respondents' pre‑treatment demographics (age, gender, income, ideology, etc.) are recorded, and choices between candidate profiles yield the reward signal for our bandit. 

\begin{enumerate}
    \item \textbf{COVID‑19 vaccine conjoint \citep{DVN/6BSJYP_2020}.} 
    \label{dataset1}
    1,970 American respondents completed a five‑task choice-based conjoint survey in July 2020, comparing two hypothetical COVID‑19 vaccines described by seven randomized attributes: efficacy, duration of protection, major side‑effect rate, minor side effects, FDA approval status, country of origin, and endorser \citep{10.1001/jamanetworkopen.2020.25594}. We flatten each respondent's demographic vector and the difference between the two vaccine attribute vectors into user–vaccine feature contexts for LinUCB.

    \item \textbf{Immigration attitudes conjoint \citep{DVN/25505_2014}.} 
    \label{dataset2}
    1,714 American adults each completed five pairwise choice tasks, selecting which of two hypothetical immigrant applicants they would admit \citep{hainmueller2015hidden}. Each immigrant profile was described by nine randomized attributes: education, profession, years of training/experience, reason for migrating, English‑language ability, prior U.S. trips, legal entry status, country of origin, and the local industry's percent foreign‑born workers. As before, we concatenate one‑hot demographics with the difference in attribute vectors to form user-choice features.

    \item \textbf{Leisure travel conjoint \citep{DVN/KA7DLE_2023}.} 
    \label{dataset3}
    In this dataset, roughly 2,100 American adults evaluated ten choice tasks, choosing between three U.S.-based leisure-travel destinations \citep{articletravelback}. Destinations are described by six randomized attributes: average July temperature, travel time, attractions, presidential election outcome of the state, recent news coverage, and community sentiments. We reduce each three‑way decision to a binary comparison by randomly selecting one of the two unchosen destinations to compare against the chosen destination, resulting in \(K=2\) per round. 
    We additionally evaluated the full three‑arm setting (\(K=3\)) and found that the regret curves differed by less than 3 percentage points. We flatten each respondent's demographics and these chosen‑vs‑unchosen attribute differences into user–destination feature vectors. 
    
\end{enumerate}

\subsection{Problem Formulation}
\label{problem formulation}

We set up the problem following the ``jump-start'' formalized by \citet{alamdari2024jumpstartingbanditsllmgenerated}. Each conjoint survey is cast as a \emph{sleeping} contextual bandit \citep{kanade2009sleeping} over \(T\) rounds. 

\begin{enumerate}
  \item \textbf{Rounds \& Arms.}  
    At round \(t\in\{1,\dots,T\}\), a subset of arms \(\mathcal{A}_t\) is presented (e.g., the two vaccines in Dataset \ref{dataset1}).

    \item \textbf{Context–Arm Features.}  
    We embed each respondent's one‑hot demographics \(u_t\) and the (chosen vs. unchosen) differences of arm attributes into a joint feature vector 
    \[
      x_{t,a} = \psi(u_t, a)\in\mathbb{R}^d.
    \]

  \item \textbf{Linear Reward Model.}
    Following standard LinUCB assumptions \citep{li2010contextual}, we assume:
    \[
      \mathbb{E}[r_t \mid x_{t,a}]
      = \theta_*^\top x_{t,a},
      \quad
      \theta_*\in\mathbb{R}^d\text{ unknown}.
    \]

  \item \textbf{Action Selection (LinUCB).}  
    At each round, choose
    \[
      a_t
      = \arg\max_{a\in\mathcal{A}_t}
      \Bigl[\hat\theta_{t-1}^\top x_{t,a}
      + \alpha\,\sqrt{x_{t,a}^\top A_{t-1}^{-1}x_{t,a}}\Bigr],
    \]
    updating
    \(
      A_t = A_{t-1} + x_{t,a_t}x_{t,a_t}^\top,
      \;b_t = b_{t-1} + r_t\,x_{t,a_t},
    \)
    as in \citet{li2010contextual, chu2011contextual}.

  \item \textbf{Regret.}  
    At round \(t\), let \(a_t\) be the arm chosen by LinUCB and  
    \[
       a_t^* \;=\;\arg\max_{a\in\mathcal{A}_t}r_t(a)
    \]
    The arm with the highest realized reward among those available.  The instantaneous regret is the random variable
    \[
      \Delta_t \;=\; r_t\bigl(a_t^*\bigr)\;-\;r_t(a_t)\;\in\{0,1\}.
    \]
    The random cumulative regret after \(T\) rounds is
    \[
      \widehat R(T)
      \;=\;\sum_{t=1}^T \Delta_t.
    \]
    In experiments we plot or report one realization of \(\widehat{R}(T)\): its trial-average over \(G=10\) independent seeds.  For theoretical comparison we refer to the expected (pseudo-)regret
    \[
      R(T)\;=\;\mathbb{E}\bigl[\widehat R(T)\bigr],
    \]
    which is a scalar quantity bounded by \(\widetilde O(\sqrt{T\,d})\) for LinUCB \citep{li2010contextual}. 
       
    \item \textbf{Ordinal Rewards.}  
    The LLM is only ever asked to compare two arms, so its outputs yield a pairwise preference rather than an absolute score.  As shown in \citet{alamdari2024jumpstartingbanditsllmgenerated}, summing these binary comparisons recovers the correct ranking of arms by success probability, even though the true reward magnitudes are not preserved.

\end{enumerate}

Note that if in Step 1, $\mathcal{A}_t$ contains all arms for all times $t$, then the problem is a classic linear contextual bandit and not a sleeping bandit.

\subsection{CBLI ``Jump-Start'' Pipeline}
\label{jump start pipeline}

We implement the ``jump‑start'' pipeline introduced in \citet{alamdari2024jumpstartingbanditsllmgenerated}:
\begin{enumerate}
  \item \textbf{Pre‑training on LLM‑Generated Priors.} 
    Generate \(N\) synthetic context–reward pairs via LLM prompts. Fit LinUCB to these to obtain warm start parameters \(\{A_a^{\mathrm{pre}},b_a^{\mathrm{pre}}\}_{a=1}^K\) 

  \item \textbf{Warm‑Start Fine‑Tuning.}  
    Initialize LinUCB with \(\{A_a=A_a^{\mathrm{pre}},\,b_a=b_a^{\mathrm{pre}}\}\) and run for \(T\) rounds on the \emph{real} conjoint data. At each round \(t\), only the arms displayed in that task are active; select via the upper‑confidence bound and update on the chosen arm.

  \item \textbf{Cold‑Start Baseline.}  
    Repeat the \(T\)‑round LinUCB procedure on the real data from scratch (\(A_a=I_d,\;b_a=0_d\)) under the same sleeping‑bandit constraints to establish a regret baseline.
\end{enumerate}

\begin{figure}
    \centering
    \includegraphics[width=0.75\linewidth]{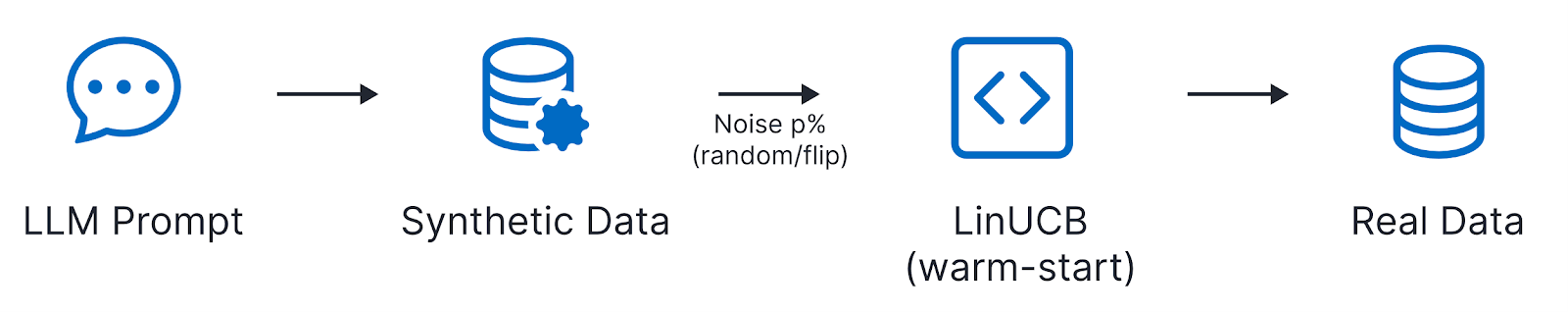}
    \caption{Overview of the CBLI evaluation framework (Noisy-CBLI). An LLM generates synthetic preference data, which is optionally corrupted with random or label-flipping noise at rate \(p\), and used to warm-start a LinUCB bandit that is then fine-tuned on real user data.}
    \label{fig:noisy-cbli-diagram}
\end{figure}

\subsection{Noise Injection Strategies}
\label{sec:noise_injection}

To evaluate CBLI’s robustness when synthetic priors are imperfect, we corrupt the LLM‐generated pre‐training labels with two controlled noise schemes, yielding what we refer to as noisy CBLI variants (\ref{fig:noisy-cbli-diagram}). Let $p$ denote the corruption rate (the proportion of synthetic samples to modify). In practical recommender systems, these two schemes correspond to two common failure modes: uninformative feedback and systematic bias. We model uninformative feedback via random response replacement, and systematic bias via preference flipping.

\begin{enumerate}
  \item \textbf{Random Response Replacement.}
    We uniformly at random select a proportion $p$ of the LLM‐generated labels (each an arm index $a\in\{1,\dots,K\}$) and overwrite each with a new arm drawn uniformly from $\{1,\dots,K\}$.  
    This simulates uninformative or arbitrary LLM mistakes.  At $p=0$ labels remain intact; at $p=1$ the entire pre‐training set is random.

  \item \textbf{Preference Flipping.}
    We randomly choose a proportion $p$ of the synthetic records and invert the original arm choice.  
    For $K=2$, flipping swaps ``A'' to ``B'' (and vice versa).  
    For $K>2$, we flip by cycling the chosen arm (e.g. \ $a \mapsto (a \bmod K) + 1$) or by selecting the least‐preferred alternative.  
    This introduces systematic bias that directly contradicts the LLM's own judgments.  At $p=1$, every label is inverted.
\end{enumerate}

Once corrupted, each noisy variant (at each noise level~$p$) replaces the original CBLI synthetic dataset. We then run the identical three‐stage pipeline from Section~\ref{jump start pipeline} on every corrupted prior to measure the impact of noise on cumulative regret. In practical recommender systems, these noise models simulate common failure modes such as uninformative feedback and systematic bias, allowing practitioners to gauge how much imperfection in LLM‑derived priors can be tolerated before online exploration must take precedence.

\subsection{Experimental Protocol and Evaluation}
\label{experiment protocol and evals}

All variants, cold‑start LinUCB and CBLI warm‑start under each noise scheme and level, are run for \(G=10\) independent trials. 
At each trial, we execute \(T\) rounds of LinUCB on the real conjoint data (Datasets 1-3) under the sleeping‑bandit constraint \(\mathcal{A}_t\).

\paragraph{Cumulative Regret.}  
We measure performance by cumulative regret $R(T)$, taking the reward $r_t \in \{0, 1\}$ to be $1$ when the bandit chooses the correct arm. For each variant, we report the trial‑average regret \(\tfrac{1}{R}\sum_{i=1}^R R_i(T)\) and its 95\% confidence interval.

\paragraph{Noise Sweep.}  
For each injection strategy (random replacement, preference flipping) and corruption rate \(p\in\{0.0,0.1,\ldots,0.7\}\), we pre‑train LinUCB on the noisy synthetic priors and then fine‑tune on the real data.  We plot regret curves up to \(T\) for each \(p\), comparing warm‑ vs. cold‑start.

\section{Theoretical Analysis}
\label{sec:main body theory}

In this section we analyze the effect of noisy LLM-generated priors on the performance of LinUCB.  
We first formalize the warm-start prior induced by Noisy–CBLI, then derive a prior-centered
confidence bound in which all pretraining effects enter through a single scalar
$\mathcal{B}_0 := \|\theta_0 - \theta^\star\|_{A_0}$.  We then make $\mathcal{B}_0$ explicit under
preference-flipping noise and target misalignment, and use this to characterize when a warm-start
is provably beneficial relative to cold-start.

\subsection{Theoretical Problem Setup and Assumptions}
\label{sec:theory_setup}

We work in the sleeping linear contextual bandit setting described in
Section~\ref{problem formulation}.  At round $t$ an availability set $\mathcal{A}_t$ and feature
vectors $\{x_{t,a}\}_{a\in \mathcal{A}_t}$ are revealed; the learner chooses $a_t\in \mathcal{A}_t$ and observes
only the reward $r_t := r_t(a_t)$.

% note to slef  --> use cf. , math ppl will know it's confer. otherwise can write explicit, @ Kevin maybe ? do I need the assumptions here?
We impose the following standard assumptions: % (cf.\ Section~\ref{math_assumptions})

\begin{itemize}
  \item \textbf{Linear realizability.}
  There exists $\theta^\star\in\mathbb{R}^d$ such that
  $\mathbb{E}[r_t(a)\mid \mathcal{F}_{t-1}, x_{t,a}] = x_{t,a}^\top\theta^\star$
  for all $t$ and $a\in A_t$.

  \item \textbf{Bounded features.}
  $\|x_{t,a}\|_2 \le 1$ for all $t$ and $a\in A_t$ (without loss of generality after rescaling).

  \item \textbf{Sub-Gaussian rewards.}
  Let $\xi_t := r_t(a_t) - x_{t,a_t}^\top \theta^\star$.
  We assume a martingale-difference and conditional sub-Gaussian condition:
  $\mathbb{E}[\xi_t \mid \mathcal{F}_{t-1}, \{x_{t,a}\}_{a\in A_t}] = 0$ and
  $\mathbb{E}[\exp(\lambda \xi_t)\mid \mathcal{F}_{t-1}, \{x_{t,a}\}_{a\in A_t}]
   \le \exp(\lambda^2\sigma^2/2)$ for all $\lambda\in\mathbb{R}$, for some $\sigma>0$.
\end{itemize}

Regret is measured against the best available arm at each round:
$a_t^\star \in \arg\max_{a\in A_t} x_{t,a}^\top \theta^\star$,
$r_t^\star := x_{t,a_t^\star}^\top \theta^\star$,
and instantaneous regret $r_t^\star - x_{t,a_t}^\top \theta^\star$.

\subsection{Noisy–CBLI Warm-Start Prior}
\label{sec:theory_warmstart}

The Noisy–CBLI warm-start uses an LLM to generate a synthetic conjoint dataset and fits a ridge
regression prior to the resulting noisy labels.

\textbf{Synthetic Responses and flip noise.}
Let $X\in\mathbb{R}^{n_s\times d}$ denote the synthetic design matrix and $y = X\theta^\star$ the corresponding ``clean'' mean success probabilities.
Let $L\in\{0,1\}^{n_s}$ be Bernoulli labels with $\mathbb{E}[L\mid X]=y$. We inject preference-flipping noise by independently drawing $F_i\sim\mathrm{Bernoulli}(p)$ for some $p \in [0, 1]$

$$\tilde L_i := (1-F_i)L_i + F_i(1-L_i).$$

Then
$\mathbb{E}[\tilde L\mid X]=(1-2p)\,y + p\,\mathbf{1}$, where
$\mathbf{1}\in\mathbb{R}^{n_s}$ is the all-ones vector.  We work with the regression proxy

\begin{equation}
\label{eq:tilde_y_model}
\tilde  y := (1-2p)\,X\theta^\star + p\,\mathbf{1} + \varepsilon,
\end{equation}

where $\varepsilon := \tilde L - \mathbb{E}[\tilde L\mid X]$ has mean zero
and conditionally $\sigma_s^2$-sub-Gaussian components by Hoeffding’s lemma.
All guarantees are stated for $p<\tfrac{1}{2}$; if an empirical flip rate
$\hat p \ge \tfrac{1}{2}$ arises, one can recode labels via effective rate
$p_{\mathrm{eff}} := \min\{\hat p,1-\hat p\}$ and apply the bounds with $p_{\mathrm{eff}} < \tfrac{1}{2}$.

\textbf{Ridge warm-start and prior error.}

Given the synthetic design matrix $X$ and the regression proxy $\tilde y$ in \eqref{eq:tilde_y_model}, we construct a ridge prior
\begin{equation}
\label{eq:ridge_prior_main}
A_0 := X^\top X + \tau_{\mathrm{pre}} I,\qquad
b_0 := X^\top \tilde y,\qquad
\theta_0 := A_0^{-1} b_0.
\end{equation}
We write $M := A_0^{-1} X^\top X$ for the corresponding shrinkage operator and define the prior mis-specification in the $A_0$-geometry as
$$
\mathcal{B}_0 := \|\theta_0 - \theta^\star\|_{A_0}
     := \sqrt{(\theta_0 - \theta^\star)^\top A_0 (\theta_0 - \theta^\star)}.
$$

At deployment time, the warm-started LinUCB algorithm initializes
$V_0 = A_0$, $\hat\theta_0 = \theta_0$ and then updates on the real conjoint
bandit stream.  The cold-start baseline instead uses
$A_0 = I$, $b_0 = 0$ (so $V_0 = I$, $\hat\theta_0=0$).

\subsection{Prior-Centered Confidence Bounds}
\label{sec:prior_centered}

We first show that the estimation error of warm-started LinUCB admits a confidence bound that is
centered at the ridge prior and depends on pretraining only through $\mathcal{B}_0$.

Let: $$ V_t := A_0 + \sum_{s\le t} x_{s,a_s} x_{s,a_s}^\top \quad (\text{so } V_t\succeq A_0)
$$

\begin{thm}[Prior-centered confidence inequality]
\label{thm:prior_centered}
For any $\delta\in(0,1)$, with probability at least $1-\delta$ the warm-started
ridge estimator satisfies, for all $t\ge 0$,
\begin{equation}
\label{eq:prior_centered}
\|\hat\theta_t - \theta^\star\|_{V_t}
\le
\beta_t(\delta) + \mathcal{B}_0,
\end{equation}
where
$$
\beta_t(\delta) := \sigma \sqrt{2\log\frac{\det(V_t)^{1/2}}{\det(A_0)^{1/2}\delta}}
$$
is the usual self-normalized variance term.
\end{thm}

The proof, Given in Appendix~\ref{app:prior_centered} follows the decomposition:

$$
V_t(\hat\theta_t - \theta^\star)
 = A_0(\theta_0 - \theta^\star)
   + \sum_{s\le t}\xi_s x_{s,a_s},
$$

bounding the first term by $\mathcal{B}_0$ in the $A_0$-norm and the second term by the
self-normalized martingale inequality of \citet{NIPS2011_e1d5be1c}.

Theorem~\ref{thm:prior_centered} induces a reward-confidence bound: for any context $x$,
\begin{equation}
\label{eq:reward_confidence_main}
|x^\top(\hat\theta_{t-1} - \theta^\star)|
\le
(\beta_{t-1}(\delta) + \mathcal{B}_0)\sqrt{x^\top V_{t-1}^{-1}x},
\end{equation}
so choosing
\begin{equation}
\label{eq:alpha_choice_main}
\alpha_t \ge \beta_{t-1}(\delta) + \mathcal{B}_0
\end{equation}
ensures that the LinUCB score is optimistic with high probability. Pretraining influences the UCB confidence radius primarily through the prior-error term $\mathcal{B}_0$, with an additional but comparatively mild logarithmic dependence on the design matrix $A_0$ inside $\beta_t(\delta)$. In practice, $\mathcal{B}_0$ is the dominant quantity governing whether warm-start improves or degrades regret.

\subsection{Flip-Noise Bias and Misalignment}
\label{sec:flip_bias_main}

We next make $\mathcal{B}_0$ explicit under preference flips and target misalignment.

Substituting \eqref{eq:tilde_y_model} into \eqref{eq:ridge_prior_main} gives
\begin{align}
\theta_0
&= A_0^{-1}X^\top\tilde y
 = (1-2p)A_0^{-1}X^\top X\theta^\star
   + pA_0^{-1}X^\top\mathbf{1}
   + A_0^{-1}X^\top\varepsilon \\
&= (1-2p)M\theta^\star
   + pA_0^{-1}X^\top\mathbf{1}
   + A_0^{-1}X^\top\varepsilon,
\end{align}
so that
\begin{equation}
\label{eq:theta0_minus_theta_main}
\theta_0-\theta^\star
= ((1-2p)M - I)\theta^\star
+ pA_0^{-1}X^\top\mathbf{1}
+ A_0^{-1}X^\top\varepsilon.
\end{equation}
Taking the $A_0$-norm and expectation over the pretraining noise $\varepsilon$ yields a bias–variance decomposition (Appendix~\ref{app:bias_decomp}):
\begin{equation}
\label{eq:B0_expectation_main}
\mathbb{E}[\mathcal{B}_0^2]
\le
\|((1-2p)M-I)\theta^\star + pA_0^{-1}X^\top\mathbf{1}\|_{A_0}^2
+ \sigma_s^2\,\mathrm{tr}(X A_0^{-1} X^\top).
\end{equation}

To interpret the flip-bias term, we diagonalize the synthetic design. Let $X^\top X = U\Lambda U^\top$ with eigenvalues $\lambda_i$ and rotated parameter $\theta^\star = U\theta^\star_U$. Because $A_0$ and $M$ share this eigenbasis,

$$
A_0 = U(\Lambda+\tau_{\mathrm{pre}}I)U^\top,
\qquad
M = U\,\mathrm{diag}\!\left(\frac{\lambda_i}{\lambda_i+\tau_{\mathrm{pre}}}\right)U^\top.
$$
This also gives the direction-wise form
\begin{equation}
\label{eq:deterministic_eig_main}
\|((1-2p)M-I)\theta^\star\|_{A_0}^2
= \sum_{i=1}^d
   \frac{(\tau_{\mathrm{pre}} + 2p\lambda_i)^2}{\lambda_i+\tau_{\mathrm{pre}}}
   (\theta^\star_{U,i})^2.
\end{equation}
In high-coverage directions where $\lambda_i\gg\tau_{\mathrm{pre}}$, this simplifies to
$\frac{(\tau_{\mathrm{pre}}+2p\lambda_i)^2}{\lambda_i+\tau_{\mathrm{pre}}}\approx 4p^2\lambda_i$, so
\begin{equation}
\label{eq:bias_approx_main}
\mathbb{E}[\mathcal{B}_0^2]
\approx
4p^2 \|(X^\top X)^{1/2}\theta^\star\|_2^2
+ \sigma_s^2\,\mathrm{tr}(X A_0^{-1} X^\top),
\end{equation}
showing that flip-induced bias grows roughly linearly in $p$ (in norm) and is amplified in directions with strong synthetic coverage.

To capture systematic misalignment between LLM-simulated and real preferences, we also consider a target shift
$$
\theta^\star_{\mathrm{syn}} = \theta^\star_{\mathrm{real}} + \Delta,
$$

where $\theta^\star_{\mathrm{syn}}$ fits the synthetic labels and $\theta^\star_{\mathrm{real}}$ fits the human data. Repeating the above with $\theta^\star_{\mathrm{syn}}$ in place of $\theta^\star$ yields, at $p=0$,
\begin{equation}
\label{eq:p0_misalignment_main}
\theta_0 - \theta^\star_{\mathrm{real}}
= (M-I)\theta^\star_{\mathrm{real}} + M\Delta,
\end{equation}
so that a large misalignment vector $\Delta$ in well-covered directions can make $\mathcal{B}_0$ large even with no injected flip noise, leading to negative transfer at $p=0$.

\subsection{When Warm-Start is Beneficial}
\label{sec:warm_help_main}

Combining the prior-centered confidence bound \eqref{eq:prior_centered} with standard LinUCB analysis yields a regret bound of the form
$$
R_{\mathrm{warm}}(T)
\; \lesssim \;
(\beta_T(\delta) + \mathcal{B}_0)\sqrt{T d \log(\cdot)},
$$
up to logarithmic factors. For the cold-start baseline with $A_0=I$, $\theta_0=0$,
the analogous bound has $\mathcal{B}_0^{\mathrm{cold}} = \|\theta^\star\|_2$ and a variance term of the same order.

Thus, a sufficient condition for CBLI warm-start to improve over cold-start is that the noisy prior satisfies
$$
\mathcal{B}_0^{\mathrm{warm}} <\mathcal{B}_0^{\mathrm{cold}},
$$

so that the reduction in variance (larger $A_0$) is not outweighed by systematic bias. Under the flip-noise and misalignment models above, \eqref{eq:B0_expectation_main} to \eqref{eq:bias_approx_main} imply that:
\begin{itemize}
    \item on aligned tasks ($\Delta\approx 0$), $\mathcal{B}_0^{\mathrm{warm}}$ increases roughly linearly with $p$, yielding a corruption threshold $p^\star$ beyond which the warm-start bound becomes worse than cold-start; and
    \item on misaligned tasks (large $\Delta$), $\mathcal{B}_0^{\mathrm{warm}}$ can exceed $\mathcal{B}_0^{\mathrm{cold}}$ already at $p=0$, explaining the empirically observed 0\%-noise failures.
\end{itemize}

Section~\ref{sec:results_cross} shows that empirical alignment measures track this theoretical picture: regimes with small estimated prior error exhibit robust warm-start gains, whereas regimes with large prior error see warm-started bandits underperform their cold-start counterparts.
\section{Empirical Results \& Discussion}
\label{sec: results and discussion}

\subsection{Preference-Flipping Noise on the COVID-19 Vaccine Conjoint Dataset}
\label{sec:results_flipping_vaccine}

\begin{figure}
    \centering
    \includegraphics[width=0.5\linewidth]{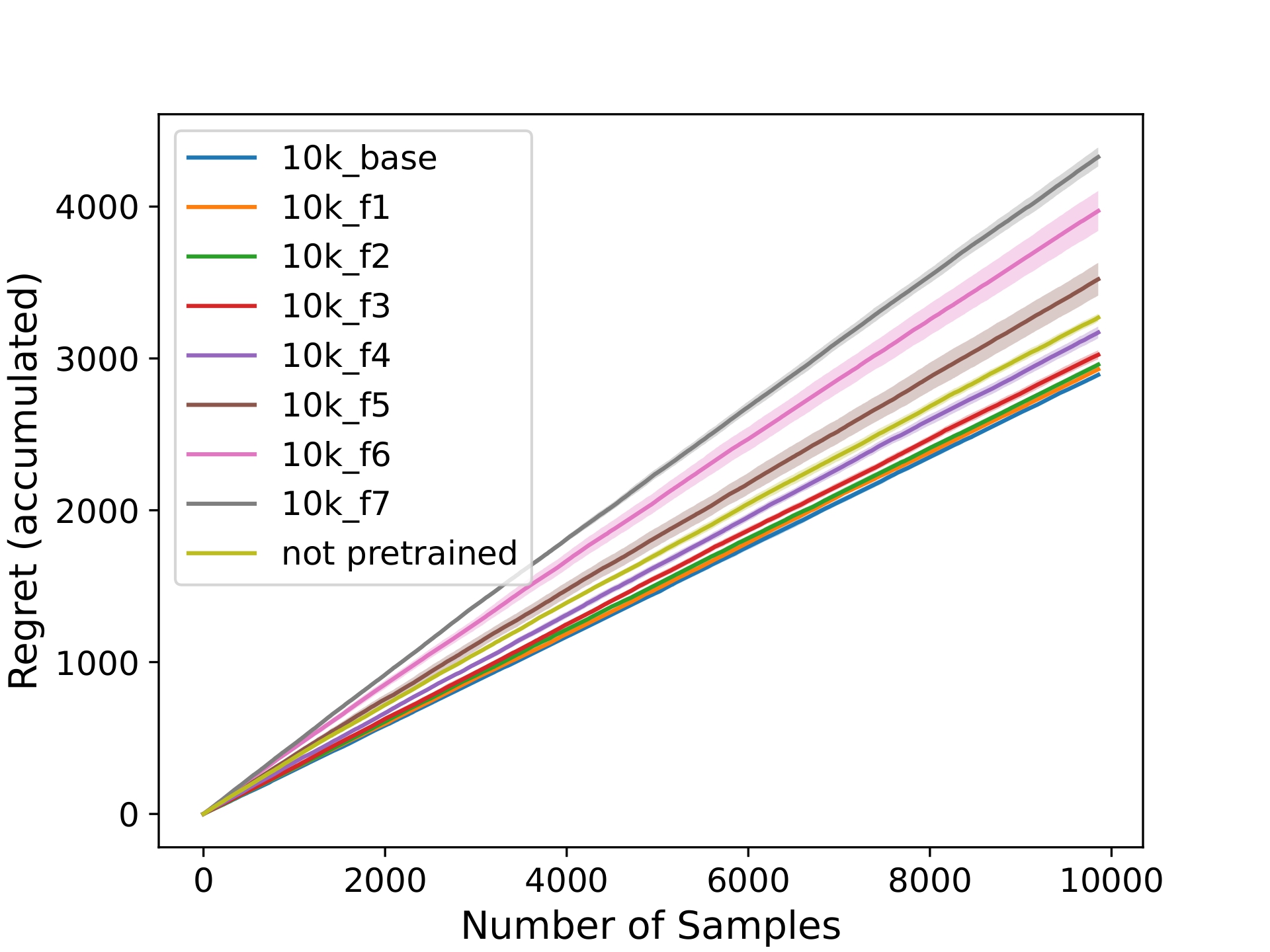}
    \caption{Cumulative regret on the COVID-19 Vaccine dataset under preference-flipping noise.  “10k\_fX” indicates X times 10\% of LLM-generated labels flipped.  Shaded regions are 95 percent CI over \(G=10\) runs.}
    \label{fig:vaccine_flipping_gpt4o}
\end{figure}

% \ref{fig:vaccine_flipping_gpt4o}
%\ref{tab:vaccine_noisy_cbli} 
Figure~\ref{fig:vaccine_flipping_gpt4o} plots the mean cumulative regret of LinUCB warm-started
on GPT-4o priors corrupted by systematic preference flipping at seven noise levels
($p \in \{0.0, 0.1, \ldots, 0.7\}$),  together with the uncorrupted baseline (``10k\_base'') and a cold-start baseline (``Not Pretrained''). Each curve is
averaged over $G=10$ trials, with shaded bands showing the 95\% confidence interval. Table~\ref{tab:gpt4o_covid_results} reports the percentage reduction in cumulative regret relative
to cold-start at horizon $T$ for different synthetic pre-training sizes and flipping rates.

\begin{itemize}
  \item \textbf{Zero noise ($p=0$).} With uncorrupted GPT-4o labels, warm-start achieves the lowest regret, quickly approaching optimal arm selection and substantially outperforming cold-start. Across all $N$, Table~ \ref{tab:gpt4o_covid_results} shows a consistent positive reduction in regret, with larger synthetic datasets mainly tightening confidence intervals and yielding modest additional gains.

  \item \textbf{Low to moderate noise (10--30 percent).} For small to moderate flipping rates, the
  warm-start curves remain below the cold-start baseline, and the corresponding entries in
  Table~\ref{tab:gpt4o_covid_results} remain positive. Preference-flipping at these levels shifts the
  regret curves upward but does not eliminate the advantage of pre-training: CBLI still converges
  faster than cold-start, especially for larger $N$.

  \item \textbf{High noise (40--70 percent).} Once flipping reaches higher levels, the benefit of
  pre-training disappears and eventually reverses. Around $p \approx 0.4$, the warm-start and
  cold-start curves become close and nearly indistinguishable, with the table entries clustering near zero.
  At $p \ge 0.5$, corrupted priors begin to harm performance: warm-started LinUCB exhibits higher
  regret than cold-start throughout much of the horizon, with the effect most pronounced for larger
  synthetic datasets where the mis-specified prior is more strongly enforced.
\end{itemize}

\begin{table}[h]
    \centering
    \small
    \caption{Percentage reduction in cumulative regret (\%\(\Delta\) Regret) for the COVID-19 Vaccine dataset using GPT-4o priors. Reported across three synthetic dataset sizes ($N$). Mean over $G=10$ seeds \(\pm\) 95\% CI.}
    \label{tab:gpt4o_covid_results}
    \begin{tabular}{lrrr}
        \toprule
        & \multicolumn{3}{c}{\textbf{Pre-training Size ($N$)}} \\
        \cmidrule(lr){2-4}
        \textbf{Noise ($p$)} & \textbf{1k} & \textbf{3k} & \textbf{10k} \\
        \midrule
        0\%  & 7.81 $\pm$ 1.31 & 10.91 $\pm$ 0.81 & 11.52 $\pm$ 0.75 \\
        10\%  & 6.50 $\pm$ 1.09 & 10.48 $\pm$ 1.19 & 10.41 $\pm$ 0.90 \\
        20\%  & 5.10 $\pm$ 1.50 & 8.43 $\pm$ 1.15 & 9.46 $\pm$ 0.92 \\
        30\%  & 4.19 $\pm$ 1.71 & 6.60 $\pm$ 1.11 & 7.49 $\pm$ 0.93 \\
        40\%  & 2.18 $\pm$ 1.75 & 2.88 $\pm$ 2.01 & 2.98 $\pm$ 1.62 \\
        50\%  & -3.85 $\pm$ 2.02 & -4.98 $\pm$ 2.27 & -7.78 $\pm$ 4.04 \\
        60\%  & -4.77 $\pm$ 2.88 & -10.90 $\pm$ 2.70 & -21.52 $\pm$ 4.60 \\
        70\%  & -7.94 $\pm$ 3.55 & -19.53 $\pm$ 4.18 & -32.37 $\pm$ 2.30 \\
        \bottomrule
    \end{tabular}
\end{table}

% Overall, the GPT-4o vaccine results indicate that, on an aligned task, CBLI warm-start remains
% beneficial under moderate preference-flipping noise, but becomes detrimental once a sufficiently
% large fraction of the synthetic labels contradict the LLM’s original preferences.

\subsection{Random-Response Noise on the COVID-19 Vaccine Conjoint Dataset}
\label{sec:results_random_vaccine}

\begin{figure}
    \centering
    \includegraphics[width=0.5\linewidth]{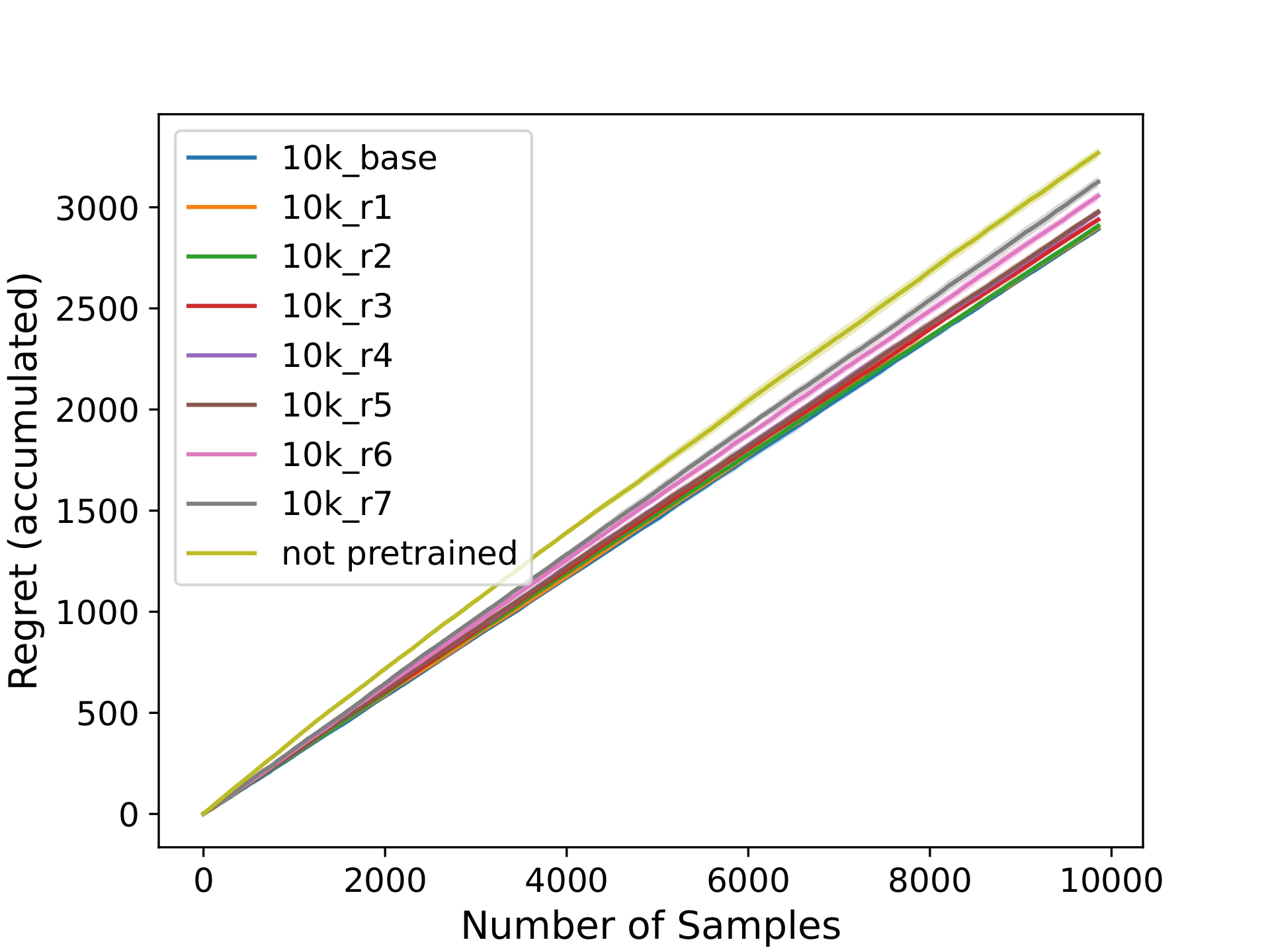}
    \caption{Cumulative regret on the COVID-19 Vaccine dataset under random-response noise.}
    \label{fig:vaccine_random_gpt4o}
\end{figure}

% \ref{fig:vaccine_random_gpt4o}
We now consider the effects of random noise on synthetic labels. Figure~\ref{fig:vaccine_random_gpt4o} plots the mean cumulative regret of LinUCB warm-started
on GPT-4o priors corrupted by random responses at the same noise levels as in
Figure~\ref{fig:vaccine_flipping_gpt4o}, together with the cold-start baseline. As before, each
curve is averaged over $G=10$ trials with 95\% confidence intervals.

\begin{itemize}
  \item \textbf{Zero noise ($p=0$).} Without corruption, warm-start again yields the lowest regret,
  converging rapidly toward optimal recommendations and reproducing the benefits observed in
  the preference-flipping setting.

  \item \textbf{Low to moderate noise (10--30 percent).} At 10--30 percent random replacements,
  the warm-start curves shift upward slightly but remain clearly below the cold-start baseline.
  CBLI continues to reduce cumulative regret relative to cold-start, indicating that a substantial
  fraction of uninformative labels can be tolerated without losing the gains from pre-training.

  \item \textbf{Moderate to high noise (40--70 percent).} For higher random-response rates, the
  warm-start curves gradually approach the cold-start curve. Unlike the preference-flipping case,
  however, we do not observe a regime where random corruption yields consistently higher regret
  than cold-start. Even at the largest tested $p$, the warm-start performance is at worst comparable
  to cold-start and often remains slightly better.
\end{itemize}

\subsection{Noise Effects Across Datasets and Models}
\label{sec:results_cross}

We next assess whether the noise--robustness patterns observed in the vaccine domain generalize across tasks and LLMs. For each dataset--model combination, we sweep random-response and preference-flipping corruption and compare warm-start to cold-start regret.

Across all aligned regimes, such as vaccine with GPT-4o or GPT-3.5, and immigration with GPT-4o or Qwen, warm-start remains beneficial under moderate corruption. Random-response noise consistently produces only gradual performance degradation and never surpasses cold-start in any of our experiments. Preference-flipping corruption induces a more structured deterioration: warm-start retains a clear advantage at low levels of flipping, but beyond a dataset-dependent threshold, the benefit disappears and eventually reverses. Although the numerical breakpoint varies, the overall pattern is consistent across models and datasets.

The travel dataset shows a qualitatively different regime. With Qwen, warm-start underperforms cold-start even at $p = 0$, and performance deteriorates monotonically under either noise type. Random-response noise does not meaningfully improve outcomes, as weakening the synthetic labels does not alter the underlying discrepancies between synthetic and human choices. Preference-flipping further accentuates these discrepancies, leading to the steepest degradation we observe. Immigration exhibits more intermediate behavior: models behave similarly under flipping, while random noise remains comparatively benign.
% \adam{Added the below.}
In addition to cross-model comparisons, we find that the strength and breakpoint clarity of warm-start gains can depend on the underlying model revision. All OpenAI results reported in this section use the Sept--Oct 2025 access window (Table~\ref{tab:model_checkpoints}). Using a GPT-3.5 Turbo synthetic prior generated from an earlier snapshot (approximately nine months prior), we observe uniformly larger regret reductions and cleaner corruption breakpoints across COVID-19, Immigration, and Travel than those reported in Table~\ref{tab:full_results_all_models}. This indicates that apparent ``alignment'' of synthetic preferences is not only task and prompt-dependent, but can drift over time even within a fixed model family. Older results can be found in Table~\ref{tab:snapshot_gpt35_flip_old}.

Together, these results demonstrate that noise sensitivity exhibits clear and reproducible structure across datasets and models, while also highlighting a separate axis of variability due to model revisions. Random-response corruption is uniformly mild, never overturning the cold-start baseline, while preference-flipping introduces directional distortions that can eliminate or reverse the warm-start benefit. Subsequent analysis in Section~\ref{sec:misalignment_results_analysis} examines how these empirical regimes relate to properties of the synthetic priors themselves.

\begin{table}[ht]
    \centering
    \small
    \caption{Percentage reduction in cumulative regret (\%$\Delta$ Regret) compared to cold-start. Reported across 4 models, 3 datasets, and 3 representative noise levels ($p \in \{0, 0.3, 0.5\}$). Mean over $G=10$ seeds.}
    \label{tab:full_results_all_models}
    \setlength{\tabcolsep}{4pt} % Adjust padding to fit page
    \begin{tabular}{lllrrr}
        \toprule
        \textbf{Model} & \textbf{Dataset} & \textbf{Noise (\%)} & \textbf{N = 1k} & \textbf{N = 3k} & \textbf{N = 10k} \\
        \midrule

        % --- GPT-3.5 Turbo ---
        \multirow{9}{*}{\textbf{GPT-3.5 Turbo}} 
            & \multirow{3}{*}{COVID-19} & 0 & 7.20 $\pm$ 1.33 & 9.06 $\pm$ 1.06 & 9.45 $\pm$ 1.17 \\
            & & 30 & 2.81 $\pm$ 3.66 & 7.06 $\pm$ 1.61 & 7.44 $\pm$ 1.34 \\
            & & 50 & -0.37 $\pm$ 4.16 & -3.76 $\pm$ 2.68 & -8.07 $\pm$ 3.45 \\
            \cmidrule{2-6}
            & \multirow{3}{*}{Immigration} & 0 & 0.91 $\pm$ 0.75 & 0.36 $\pm$ 2.18 & 0.04 $\pm$ 2.22 \\
            & & 30 & 0.40 $\pm$ 0.89 & -4.72 $\pm$ 3.38 & -2.63 $\pm$ 2.54 \\
            & & 50 & -0.22 $\pm$ 1.36 & -5.09 $\pm$ 4.41 & -4.60 $\pm$ 2.73 \\
            \cmidrule{2-6}
            & \multirow{3}{*}{Travel} & 0 & 2.45 $\pm$ 0.75 & 1.23 $\pm$ 1.02 & 2.63 $\pm$ 0.76 \\
            & & 30 & -2.17 $\pm$ 0.95 & -1.42 $\pm$ 1.02 & -1.69 $\pm$ 0.91 \\
            & & 50 & -2.71 $\pm$ 0.74 & -1.77 $\pm$ 0.76 & -2.23 $\pm$ 1.19 \\
        \midrule

        % --- GPT-4o ---
        \multirow{9}{*}{\textbf{GPT-4o}} 
            & \multirow{3}{*}{COVID-19} & 0  & 7.81 $\pm$ 1.31 & 10.91 $\pm$ 0.81 & 11.52 $\pm$ 0.75 \\
            & & 30 & 4.19 $\pm$ 1.71 & 6.60 $\pm$ 1.11 & 7.49 $\pm$ 0.93 \\
            & & 50 & -3.85 $\pm$ 2.02 & -4.98 $\pm$ 2.27 & -7.78 $\pm$ 4.04 \\
            \cmidrule{2-6}
            & \multirow{3}{*}{Immigration} & 0 & 1.75 $\pm$ 0.79 & 0.96 $\pm$ 0.81 & 0.45 $\pm$ 1.58 \\
            & & 30 & 0.52 $\pm$ 0.96 & 0.15 $\pm$ 1.32 & -2.06 $\pm$ 2.82 \\
            & & 50 & -1.35 $\pm$ 1.24 & -3.66 $\pm$ 2.14 & -4.60 $\pm$ 2.73 \\
            \cmidrule{2-6}
            & \multirow{3}{*}{Travel} & 0 & 2.83 $\pm$ 0.50 & 1.69 $\pm$ 0.61 & 3.78 $\pm$ 0.91 \\
            & & 30 & -1.19 $\pm$ 0.76 & 1.44 $\pm$ 1.24 & 2.21 $\pm$ 0.95 \\
            & & 50 & -1.58 $\pm$ 1.12 & 1.61 $\pm$ 0.86 & 1.12 $\pm$ 1.10 \\
        \midrule

        % --- Llama 3.1 ---
        \multirow{9}{*}{\textbf{Llama 3.1}} 
            & \multirow{3}{*}{COVID-19} & 0 & 5.57 $\pm$ 1.34 & 10.41 $\pm$ 1.01 & 10.57 $\pm$ 0.96 \\
            & & 30 & 3.99 $\pm$ 2.54 & 4.99 $\pm$ 1.22 & 5.38 $\pm$ 1.92 \\
            & & 50 & -1.79 $\pm$ 2.53 & -2.24 $\pm$ 2.46 & -5.32 $\pm$ 3.06 \\
            \cmidrule{2-6}
            & \multirow{3}{*}{Immigration} & 0 & 0.88 $\pm$ 1.79 & -0.99 $\pm$ 2.06 & -1.40 $\pm$ 2.86 \\
            & & 30 & 0.53 $\pm$ 2.44 & -3.94 $\pm$ 3.78 & -3.81 $\pm$ 4.63 \\
            & & 50 & -0.90 $\pm$ 2.06 & -2.98 $\pm$ 5.08 & -3.31 $\pm$ 6.29 \\
            \cmidrule{2-6}
            & \multirow{3}{*}{Travel} & 0 & -1.29 $\pm$ 1.14 & -1.44 $\pm$ 1.88 & 0.80 $\pm$ 0.89 \\
            & & 30 & -2.67 $\pm$ 1.53 & 2.85 $\pm$ 2.45 & -3.63 $\pm$ 2.24 \\
            & & 50 & -3.89 $\pm$ 2.29 & 3.23 $\pm$ 2.44 & -7.19 $\pm$ 1.35 \\
        \midrule

        % --- Qwen 3 ---
        \multirow{9}{*}{\textbf{Qwen 3}} 
            & \multirow{3}{*}{COVID-19} & 0 & 3.13 $\pm$ 1.87 & 5.59 $\pm$ 1.37 & 7.08 $\pm$ 1.15 \\
            & & 30 & 1.27 $\pm$ 1.80 & 2.17 $\pm$ 2.54 & -1.34 $\pm$ 1.61 \\
            & & 50 & -1.76 $\pm$ 2.17 & -2.48 $\pm$ 1.81 & -10.40 $\pm$ 2.77 \\
            \cmidrule{2-6}
            & \multirow{3}{*}{Immigration} & 0 & 1.63 $\pm$ 1.84 & 0.61 $\pm$ 2.11 & 0.18 $\pm$ 2.13 \\
            & & 30 & 0.54 $\pm$ 1.36 & -2.42 $\pm$ 4.99 & -2.65 $\pm$ 3.20 \\
            & & 50 & -2.01 $\pm$ 1.45 & -4.26 $\pm$ 2.30 & -4.78 $\pm$ 6.02 \\
            \cmidrule{2-6}
            & \multirow{3}{*}{Travel} & 0 & -1.44 $\pm$ 0.96 & -2.36 $\pm$ 1.87 & -2.52 $\pm$ 2.56 \\
            & & 30 & -0.91 $\pm$ 1.05 & -2.93 $\pm$ 2.64 & -2.10 $\pm$ 2.31 \\
            & & 50 & -2.27 $\pm$ 1.08 & -2.07 $\pm$ 1.33 & -0.62 $\pm$ 1.27 \\
        \bottomrule
    \end{tabular}
\end{table}

\subsection{misalignment analysis}
\label{sec:misalignment_results_analysis}

Our theory identifies the prior-error term
$$
\mathcal{B}_0 \;=\; \bigl\lVert \theta_0 - \theta^\star \bigr\rVert_{A_0},
\qquad 
A_0 = X_{\mathrm{syn}}^\top X_{\mathrm{syn}} + \tau I,
$$
as the central quantity governing whether an LLM-generated warm start helps or harms LinUCB. In the prior-centered confidence bound of Section~\ref{sec:main body theory}, the exploration radius satisfies

$$
\lVert \hat\theta_t - \theta^\star \rVert_{V_t}
\;\le\;
\beta_t(\delta) + \mathcal{B}_0,
$$
where $\beta_t(\delta)$ depends on $A_0$ only through a mild logarithmic term, while $\mathcal{B}_0$ enters additively and therefore dominates the warm-start effect. To connect this analysis to practice, we estimate $\theta_0$ by ridge regression on the synthetic LLM responses. Since the real parameter $\theta^\star$ is unobserved, we approximate it using a ridge fit on the real conjoint dataset. We compute: 
$$
\widehat{\mathcal{B}}_0 \;=\; \lVert \theta_0 - \theta_{\mathrm{real}} \rVert_{A_0}.
$$
In the shared feature space used by the bandit.

Table~\ref{tab:prior_error} reports $\widehat{\mathcal{B}}_0$ for each model-dataset pair (and can vary across model snapshots). Across all three datasets, at $N=10\text{k}$ rounds and $p=0$ (Table~\ref{tab:full_results_all_models}), the ordering of models by regret reduction relative to cold-start matches the ordering by $\widehat{\mathcal{B}}_0$. On the COVID-19 (vaccine) dataset, GPT-4o has the smallest prior error ($\widehat{\mathcal{B}}_0 \approx 44.4$) and achieves the largest regret reduction ($\approx 11.5\%$), while Qwen has the largest prior error ($\widehat{\mathcal{B}}_0 \approx 62.3$) and the smallest improvement. On Immigration, GPT-4o again has the smallest $\widehat{\mathcal{B}}_0$ ($\approx 56.2$) and the best warm-start performance, whereas Llama has the largest prior error ($\approx 59.2$) and is the only model that exhibits negative transfer at $p=0$. Finally, on Travel, GPT-4o has the smallest prior error ($\widehat B_0 \approx 26.9$) and the largest regret reduction ($\approx 3.8\%$), while Qwen has the largest prior error ($\approx 28.9$) and the strongest negative transfer ($\approx -2.5\%$). Thus, within each dataset, a smaller estimated prior error is consistently associated with more beneficial warm start.

\begin{table}[t]
    \centering
    \small
    \caption{Estimated prior error $\widehat{\mathcal{B}}_0 = \|\theta_0 - \theta_{\mathrm{real}}\|_{A_0}$ for each model and dataset. Lower values indicate closer alignment between LLM-synthetic and human preferences in the $A_0$-geometry.}
    \label{tab:prior_error}
    \begin{tabular}{lccc}
        \toprule
        \textbf{Model} & \textbf{COVID-19} & \textbf{Immigration} & \textbf{Travel} \\
        \midrule
        GPT-3.5 Turbo & 51.0 & 57.1 & 28.1 \\
        GPT-4o        & 44.4 & 56.2 & 26.9 \\
        Llama 3.1     & 46.2 & 59.2 & 28.2 \\
        Qwen 3        & 62.3 & 56.9 & 28.9 \\
        \bottomrule
    \end{tabular}
\end{table}

We note that the absolute scale of $\widehat{\mathcal{B}}_0$ varies across datasets, reflecting differences in the synthetic design $X_{\mathrm{syn}}$ and the $A_0$-geometry. Our sufficient condition from Section~\ref{sec:warm_help_main} compares the warm-start prior error to the cold-start baseline, $\widehat{\mathcal{B}}_0^{\mathrm{cold}} = \|\theta^\star\|_2$, which we do not directly estimate here. Instead, we use $\widehat{\mathcal{B}}_0$ as a task-specific \emph{risk score} and interpret it through a simple within-dataset decision rule: warm starts with sufficiently small $\widehat{\mathcal{B}}_0$ are predicted to be beneficial, while larger values are predicted to yield marginal gains or negative transfer. Empirically, within each dataset, models with smaller $\widehat{\mathcal{B}}_0$ reliably improve over cold-start, whereas models with larger $\widehat{\mathcal{B}}_0$ yield marginal gains or negative transfer. This supports the view of $B_0$ as a useful (though not perfectly calibrated) diagnostic for when LLM-derived priors are likely to help or hurt contextual-bandit performance.

\section{Conclusion}
\label{sec:conclusion}

We examined how noise and underlying preference mismatch affect the usefulness of LLM-generated priors for warm-starting contextual bandits. Across three conjoint datasets and multiple LLMs, we found that warm-start improves regret only when synthetic preferences track human choices closely. In these aligned settings, random-response corruption is uniformly mild, and warm-start remains beneficial under moderate preference-flipping noise before losing its advantage at higher corruption levels. In contrast, on misaligned tasks, warm-start can underperform cold-start even at $p = 0$, and both noise types further degrade performance.

To explain these observations, we developed a prior-centered analysis in which pretraining affects regret through a single prior-error term, $\mathcal{B}_0 = \|\theta_0 - \theta^\star\|_{A_0}$, and derived sufficient conditions under which warm-start cannot worsen regret relative to cold-start. Empirically, transitions between helpful and harmful behavior align with changes in this prior error: random noise does not increase $\mathcal{B}_0$ in harmful directions, whereas preference-flipping introduces directional biases that rapidly enlarge it. Moreover, our estimates $\widehat{\mathcal{B}}_0$ track the empirical outcomes: within each dataset, models with smaller prior error consistently achieve larger regret reductions over cold-start, while those with larger prior error yield marginal gains or negative transfer. These findings suggest that LLM-generated priors are most valuable when alignment is high and corruption is moderate, and should be deployed cautiously in settings where synthetic and real preferences may diverge.
 
\section{Limitations}
\label{sec:limitations}

Our work has provided a systematic analysis of the use of synthetic LLM priors for bandit recommender systems; however, limitations remain. The warm-start procedure depends on a fixed set of prompts, yet LLM outputs are highly prompt-sensitive: minor wording changes, added context, or altered arm order \citep{pezeshkpour2023largelanguagemodelssensitivity} can materially shift the synthetic labels, so the evaluation may provide an unduly narrow estimate of variance in the prior \citep{sclar2024quantifyinglanguagemodelssensitivity, errica2025didiwrongquantifying}. The injected noise follows an independent and identically distributed random-replacement or label-flip model, whereas empirical LLM errors exhibit heteroskedastic and context-correlated structure. Consequently, the corruption sweep may mischaracterize real-world error modes, and future work should consider context-dependent or structured noise \citep{xia2020partdependentlabelnoiseinstancedependent}. Our theoretical analysis relies on a high-coverage approximation ($\lambda_i \gg \tau_{\mathrm{pre}}$) and focuses on the upper confidence bound. In sparse regimes, the bias-variance trade-off may deviate from our quadratic scaling, and a lower bound is needed to prove failure tolerances more rigorously. Commercial LLMs are governed by evolving safety guardrails that can refuse or reshape responses about sensitive content, altering the effective reward distribution and introducing non-stationarity that violates standard regret assumptions \citep{pantha2024challengesguardrailinglargelanguage}. More broadly, model revisions can shift synthetic preference distributions over time: for GPT-3.5 Turbo, we observe materially stronger and cleaner warm-start gains from an earlier snapshot than from the Sept--Oct 2025 access window used for the main results (Appendix~\ref{app:model info}; Table~\ref{tab:snapshot_gpt35_flip_old}). Lastly, LLMs encode demographic and ideological biases from their training data. When such biases manifest in synthetic preferences \citep{wyllie2024fairnessfeedbackloopstraining}, they are inherited by the bandit and can persist downstream. These biases may not be immediately observable, so despite potential early-stage regret gains, fairness auditing and bias mitigation remain essential challenges \citep{gallegos2024biasfairnesslargelanguage}.

\section{Future Works}
\label{sec:Future_Works}

Future work should seek to derive lower regret bounds that mathematically confirm the ``tipping point'' trends recorded in our experiments. On the practical side, we envision a lightweight alignment estimator acting as a statistical pre-check to flag potentially harmful priors before they are deployed. Finally, extending the Noisy-CBLI framework beyond linear assumptions to neural bandits would allow for more sophisticated modeling of the context-dependent noise often seen in real-world LLM outputs. 

% \section{Acknowledgements}
% \label{sec:acknowledgements}
% \adam{Comment out for OpenReview Submission.}
% The authors would like to thank the reviewers and hosts of the Second Workshop on Generative AI for Recommender Systems and Personalization 2025, held in conjunction with KDD, for their helpful comments on an early version of this work.   
\section{Broader Impact Statement}
\label{sec:broader impact statement}

This work investigates the reliability of using Large Language Models to initialize recommender systems. While our primary contribution is technical, establishing robustness thresholds for synthetic priors, we identify several ethical implications regarding the deployment of this technology.

The most significant risk in Noisy-CBLI frameworks is the potential for feedback loops where LLM-encoded biases are transferred to the bandit policy. As noted in our experiments with the Immigration and Vaccine datasets, LLM priors can be opinionated. If a synthetic prior contains demographic or ideological biases (e.g., favoring specific groups in the Immigration task), the warm-started bandit will operationalize this discrimination immediately upon deployment, potentially disadvantaging real users or items before human feedback can correct the policy. Practitioners must audit synthetic priors for fairness, not just regret minimization, before initialization. 

Our evaluation includes high-stakes domains, such as public health (COVID-19 vaccination). Deploying warm-started bandits in such settings carries the risk of amplifying hallucinations or medical misinformation inherent in the LLM. Our theoretical analysis provides a safeguard against this by defining a breakdown threshold, offering practitioners principled guidelines detailing when to reject synthetic priors that do not meet strict alignment standards, thereby preventing the deployment of unreliable systems in critical contexts.

This research involved substantial computational resources for generating synthetic data and simulating bandit trajectories across multiple models (Llama, Qwen, GPT families). While the immediate cost is non-negligible, our findings suggest that LLM initialization is detrimental in high-noise or misaligned settings. This insight potentially reduces long-term environmental impact by discouraging the wasteful deployment of generative models in domains where they offer no performance benefit over cold-start algorithms. 
\bibliography{main}
\bibliographystyle{tmlr}

\appendix
% \section{Appendix}
\section{Additional Theoretical Details}
\label{app:theory}

In this appendix we provide proofs and derivations for the results stated in
Section~\ref{sec:main body theory}.  We keep the notation from the main text:
$A_0$, $\theta_0$, and $V_t$ denote the ridge pretraining precision, prior
parameter, and cumulative design matrix, respectively, and
$\mathcal{B}_0 := \|\theta_0-\theta^\star\|_{A_0}$ is the prior
mis-specification measured in the $A_0$-Mahalanobis norm.

% \vspace{0.5em}
\noindent
\textbf{Notation.}
For any symmetric positive semi definite matrix $G\in\mathbb{R}^{d\times d}$ and
vector $v\in\mathbb{R}^d$ we write
$$
\|v\|_G := \sqrt{v^\top G v},
\qquad
\langle u,v\rangle_G := u^\top G v.
$$
We write $\|\cdot\|_2$ for the Euclidean norm, and $A\succeq B$ for Loewner
order on symmetric matrices.

\subsection{Proof of Theorem~\ref{thm:prior_centered}}
\label{app:prior_centered}

Recall that the warm-started ridge estimator is defined by
$$
A_0 := X^\top X + \tau_{\text{pre}} I,\qquad
b_0 := X^\top \tilde y,\qquad
\theta_0 := A_0^{-1} b_0,
$$
and the online design matrix is
$$
V_t := A_0 + \sum_{s \le t} x_{s,a_s} x_{s,a_s}^\top.
$$
The estimator at time $t$ has the usual ridge form
$$
\hat\theta_t = V_t^{-1}\left(A_0 \theta_0 + \sum_{s \le t} r_s x_{s,a_s}\right).
$$

\paragraph{Error decomposition.}
Using $r_s = x_{s,a_s}^\top \theta^\star + \xi_s$, we write
\begin{align*}
V_t(\hat\theta_t - \theta^\star)
&= A_0\theta_0 + \sum_{s\le t} r_s x_{s,a_s}
   - \left(A_0 + \sum_{s\le t} x_{s,a_s} x_{s,a_s}^\top\right)\theta^\star \\
&= A_0(\theta_0 - \theta^\star)
   + \sum_{s\le t} \bigl(r_s - x_{s,a_s}^\top\theta^\star\bigr)x_{s,a_s} \\
&= A_0(\theta_0 - \theta^\star)
   + \sum_{s\le t} \xi_s x_{s,a_s}.
\end{align*}
Multiplying by $V_t^{-1/2}$ on the left gives
\begin{equation}
\label{eq:app_triangular_decomp}
V_t^{1/2}(\hat\theta_t - \theta^\star)
  = V_t^{-1/2}A_0(\theta_0 - \theta^\star)
  + V_t^{-1/2}\sum_{s\le t} \xi_s x_{s,a_s}.
\end{equation}
Taking Euclidean norms and applying the triangle inequality,
\begin{equation}
\label{eq:app_tri_bound}
\|\hat\theta_t - \theta^\star\|_{V_t}
  = \|V_t^{1/2}(\hat\theta_t - \theta^\star)\|_2
 \le \underbrace{\|V_t^{-1/2}A_0(\theta_0 - \theta^\star)\|_2}_{\text{prior term}}
   + \underbrace{\left\|V_t^{-1/2}\sum_{s\le t} \xi_s x_{s,a_s}\right\|_2}_{\text{noise term}}.
\end{equation}

\paragraph{Bounding the prior term by \(\mathcal{B}_0\).}
We first control the deterministic term.  Using the definition of the
$A_0$-norm and the fact that $V_t\succeq A_0$, we have
\begin{align*}
\|V_t^{-1/2}A_0(\theta_0 - \theta^\star)\|_2^2
  &= (\theta_0 - \theta^\star)^\top A_0 V_t^{-1} A_0 (\theta_0 - \theta^\star) \\
  &= \|A_0^{1/2}(\theta_0 - \theta^\star)\|_{A_0^{1/2}V_t^{-1}A_0^{1/2}}^2.
\end{align*}
Since $V_t = A_0 + \sum_{s\le t} x_{s,a_s}x_{s,a_s}^\top \succeq A_0$, we have
$V_t^{-1} \preceq A_0^{-1}$, and hence
$A_0^{1/2}V_t^{-1}A_0^{1/2} \preceq I$.  Therefore,
$$
\|V_t^{-1/2}A_0(\theta_0 - \theta^\star)\|_2^2
  \le \|A_0^{1/2}(\theta_0 - \theta^\star)\|_2^2
  = \|\theta_0 - \theta^\star\|_{A_0}^2
  = \mathcal{B}_0^2,
$$
so
\begin{equation}
\label{eq:app_prior_bound}
\|V_t^{-1/2}A_0(\theta_0 - \theta^\star)\|_2 \le \mathcal{B}_0.
\end{equation}

\paragraph{Bounding the noise term.}
The second term in \eqref{eq:app_tri_bound} is the self-normalized noise process
$$
\left\|V_t^{-1/2}\sum_{s\le t} \xi_s x_{s,a_s}\right\|_2
 = \left\|\sum_{s\le t} \xi_s x_{s,a_s}\right\|_{V_t^{-1}}.
$$
Under the sub-Gaussian noise and bounded-feature assumptions, the
self-normalized concentration inequality of \citet{NIPS2011_e1d5be1c} implies
that for any $\delta\in(0,1)$, with probability at least $1-\delta$,
\begin{equation}
\label{eq:app_self_normalized}
\left\|\sum_{s\le t} \xi_s x_{s,a_s}\right\|_{V_t^{-1}}
\le
\sigma \sqrt{2\log\frac{\det(V_t)^{1/2}}{\det(A_0)^{1/2}\delta}}
= \beta_t(\delta)
\end{equation}
simultaneously for all $t\ge 0$.

\paragraph{Combining the two terms.}
Substituting \eqref{eq:app_prior_bound} and \eqref{eq:app_self_normalized} into
\eqref{eq:app_tri_bound} yields, on the same high-probability event and for all
$t\ge 0$,
$$
\|\hat\theta_t - \theta^\star\|_{V_t}
\le
\beta_t(\delta)
+ \mathcal{B}_0,
$$
which is exactly the prior-centered confidence inequality
\eqref{eq:prior_centered}.  This proves Theorem~\ref{thm:prior_centered}.

Finally, applying \eqref{eq:prior_centered} to a fixed context $x$ gives
$$
|x^\top(\hat\theta_{t-1}-\theta^\star)|
= |\langle \hat\theta_{t-1}-\theta^\star, x\rangle|
\le \|\hat\theta_{t-1}-\theta^\star\|_{V_{t-1}}\cdot \|x\|_{V_{t-1}^{-1}}
\le (\beta_{t-1}(\delta) + \mathcal{B}_0)\sqrt{x^\top V_{t-1}^{-1}x},
$$
which is the reward-confidence bound \eqref{eq:reward_confidence_main}.

\subsection{Bias–Variance Decomposition for $\mathcal{B}_0^2$}
\label{app:bias_decomp}

We now derive the decomposition of $\mathcal{B}_0^2$ and its expectation under
the flip-noise model.  Recall the regression proxy and ridge prior
\eqref{eq:tilde_y_model}–\eqref{eq:ridge_prior_main}:
$$
\tilde y = (1-2p)X\theta^\star + p\mathbf{1} + \varepsilon,
\qquad
A_0 = X^\top X + \tau_{\mathrm{pre}} I,
\qquad
\theta_0 = A_0^{-1} X^\top \tilde y.
$$
Substituting $\tilde y$ into $\theta_0$ yields
\begin{align*}
\theta_0
&= A_0^{-1} X^\top\tilde y \\
&= A_0^{-1}X^\top\bigl((1-2p)X\theta^\star + p\mathbf{1} + \varepsilon\bigr) \\
&= (1-2p)A_0^{-1}X^\top X\theta^\star
   + pA_0^{-1}X^\top\mathbf{1}
   + A_0^{-1}X^\top\varepsilon \\
&= (1-2p)M\theta^\star
   + pA_0^{-1}X^\top\mathbf{1}
   + A_0^{-1}X^\top\varepsilon,
\end{align*}
where $M := A_0^{-1}X^\top X$.  Subtracting $\theta^\star$ and regrouping gives
\begin{equation}
\label{eq:app_theta0_minus_theta}
\theta_0 - \theta^\star
= \underbrace{((1-2p)M - I)\theta^\star + pA_0^{-1}X^\top\mathbf{1}}_{D}
+ \underbrace{A_0^{-1}X^\top\varepsilon}_{\text{pretraining noise}}.
\end{equation}
Define the deterministic component
$$
D := ((1-2p)M - I)\theta^\star + pA_0^{-1}X^\top\mathbf{1}.
$$
Then the prior error in the $A_0$-norm satisfies
\begin{align*}
\mathcal{B}_0^2
&= \|\theta_0-\theta^\star\|_{A_0}^2
 = \|D + A_0^{-1}X^\top\varepsilon\|_{A_0}^2 \\
&= \|D\|_{A_0}^2
 + \|A_0^{-1}X^\top\varepsilon\|_{A_0}^2
 + 2\langle D, A_0^{-1}X^\top\varepsilon\rangle_{A_0}.
\end{align*}
Using $\|v\|_{A_0}^2 = v^\top A_0 v$ and the definition of the $A_0$-inner product,
we can write the three terms explicitly as
\begin{align*}
\|D\|_{A_0}^2
&= D^\top A_0 D, \\
\|A_0^{-1}X^\top\varepsilon\|_{A_0}^2
&= \varepsilon^\top X A_0^{-1} A_0 A_0^{-1} X^\top\varepsilon
 = \varepsilon^\top X A_0^{-1}X^\top\varepsilon, \\
\langle D, A_0^{-1}X^\top\varepsilon\rangle_{A_0}
&= D^\top A_0 A_0^{-1}X^\top\varepsilon
 = D^\top X^\top\varepsilon.
\end{align*}

\paragraph{Taking expectation over pretraining noise.}
We now take expectation with respect to the pretraining noise
$\varepsilon$ conditional on $X$, using the assumptions
$$
\mathbb{E}[\varepsilon\mid X]=0,
\qquad
\mathbb{E}[\varepsilon\varepsilon^\top\mid X]\preceq \sigma_s^2 I.
$$
The cross term has mean zero:
$$
\mathbb{E}[\langle D, A_0^{-1}X^\top\varepsilon\rangle_{A_0}\mid X]
 = D^\top X^\top \mathbb{E}[\varepsilon\mid X]
 = 0.
$$
For the noise quadratic term we use the trace identity
$\mathbb{E}[z^\top A z] = \mathrm{tr}(A\,\mathbb{E}[zz^\top])$ to obtain
\begin{align*}
\mathbb{E}\bigl[\|A_0^{-1}X^\top\varepsilon\|_{A_0}^2 \mid X\bigr]
&= \mathbb{E}\bigl[\varepsilon^\top X A_0^{-1}X^\top\varepsilon \mid X\bigr] \\
&= \mathrm{tr}\bigl(X A_0^{-1} X^\top \mathbb{E}[\varepsilon\varepsilon^\top\mid X]\bigr) \\
&\le \sigma_s^2\,\mathrm{tr}\bigl(X A_0^{-1} X^\top\bigr).
\end{align*}
Combining these pieces yields
\begin{align*}
\mathbb{E}[\mathcal{B}_0^2 \mid X]
&\le \|D\|_{A_0}^2
  + \sigma_s^2\,\mathrm{tr}(X A_0^{-1} X^\top) \\
&= \|((1-2p)M-I)\theta^\star + pA_0^{-1}X^\top\mathbf{1}\|_{A_0}^2
  + \sigma_s^2\,\mathrm{tr}(X A_0^{-1} X^\top),
\end{align*}
which is exactly the bound stated in \eqref{eq:B0_expectation_main}.

\subsection{Eigenbasis Expansion and High-Coverage Approximation}
\label{app:eigen_expansion}

We derive the direction-wise expression \eqref{eq:deterministic_eig_main} for
the deterministic flip-bias term and its high-coverage approximation.

\paragraph{Joint diagonalization.}
Let $X^\top X = U\Lambda U^\top$ be the eigendecomposition of the synthetic
Gram matrix, with $\Lambda = \mathrm{diag}(\lambda_1,\dots,\lambda_d)$ and
$U$ orthogonal.  Because
$$
A_0 = X^\top X + \tau_{\mathrm{pre}} I
    = U(\Lambda + \tau_{\mathrm{pre}}I)U^\top,
$$
we have
$$
A_0^{-1} = U(\Lambda+\tau_{\mathrm{pre}}I)^{-1}U^\top.
$$
The shrinkage operator $M:=A_0^{-1}X^\top X$ shares the same eigenbasis:
$$
M = U\,\mathrm{diag}\!\left(\frac{\lambda_i}{\lambda_i+\tau_{\mathrm{pre}}}\right)U^\top.
$$

Write $\theta^\star = U\theta^\star_U$ in the eigenbasis.  Then
$$
((1-2p)M - I)\theta^\star
  = U\,\mathrm{diag}\!\left((1-2p)\frac{\lambda_i}{\lambda_i+\tau_{\mathrm{pre}}} - 1\right)\theta^\star_U.
$$
The diagonal entries simplify to
$$
(1-2p)\frac{\lambda_i}{\lambda_i+\tau_{\mathrm{pre}}} - 1
= -\frac{\tau_{\mathrm{pre}} + 2p\lambda_i}{\lambda_i+\tau_{\mathrm{pre}}}.
$$
Hence the $i$-th coordinate of $((1-2p)M - I)\theta^\star$ in the $U$-basis is
$$
v_i := -\frac{\tau_{\mathrm{pre}} + 2p\lambda_i}{\lambda_i+\tau_{\mathrm{pre}}}\,\theta^\star_{U,i}.
$$

\paragraph{Computing the $A_0$-norm.}
The $A_0$-norm of $((1-2p)M-I)\theta^\star$ satisfies
$$
\|((1-2p)M-I)\theta^\star\|_{A_0}^2
 = \sum_{i=1}^d (\lambda_i+\tau_{\mathrm{pre}}) v_i^2,
$$
because $A_0$ is diagonal with entries $(\lambda_i+\tau_{\mathrm{pre}})$ in the
$U$-basis.  Substituting the expression for $v_i$ gives
$$
\|((1-2p)M-I)\theta^\star\|_{A_0}^2
 = \sum_{i=1}^d
    (\lambda_i+\tau_{\mathrm{pre}})
    \left(\frac{\tau_{\mathrm{pre}} + 2p\lambda_i}{\lambda_i+\tau_{\mathrm{pre}}}\right)^2
    (\theta^\star_{U,i})^2
 = \sum_{i=1}^d
    \frac{(\tau_{\mathrm{pre}} + 2p\lambda_i)^2}{\lambda_i+\tau_{\mathrm{pre}}}
    (\theta^\star_{U,i})^2,
$$
which is exactly \eqref{eq:deterministic_eig_main}.

\paragraph{High-coverage approximation.}
In directions where the synthetic design has strong coverage,
$\lambda_i\gg\tau_{\mathrm{pre}}$, we have
$$
\frac{(\tau_{\mathrm{pre}} + 2p\lambda_i)^2}{\lambda_i+\tau_{\mathrm{pre}}}
  \approx \frac{(2p\lambda_i)^2}{\lambda_i}
  = 4p^2 \lambda_i,
$$
so
$$
\|((1-2p)M-I)\theta^\star\|_{A_0}^2
\approx 4p^2 \sum_{i=1}^d \lambda_i (\theta^\star_{U,i})^2
 = 4p^2 \|(X^\top X)^{1/2}\theta^\star\|_2^2.
$$
Since $A_0^{1/2}$ and $(X^\top X)^{1/2}$ are comparable in these directions
($\lambda_i\gg\tau_{\mathrm{pre}}$ implies $\lambda_i+\tau_{\mathrm{pre}}\approx\lambda_i$),
this yields the norm-level approximation
$$
\|((1-2p)M-I)\theta^\star\|_{A_0}
\approx 2p\,\|A_0^{1/2}\theta^\star\|,
$$
which is the heuristic form used in the main text
(cf.\ \eqref{eq:bias_approx_main}).  The exact dependence on $(p,\lambda_i,\tau_{\mathrm{pre}})$
is given by \eqref{eq:deterministic_eig_main}.

\subsection{High-Probability Control of the Pretraining-Noise Term}
\label{app:noise_hp}

The expectation bound \eqref{eq:B0_expectation_main} controls the contribution
of the pretraining noise $\varepsilon$ in $\mathbb{E}[\mathcal{B}_0^2]$.  For
completeness, we record a high-probability bound on the same quantity; this is
not used directly in the main text but may be useful for refined regret bounds.

Recall from \eqref{eq:app_theta0_minus_theta} that the noise component of the
prior error is
$$
\|A_0^{-1}X^\top\varepsilon\|_{A_0}
 = \|A_0^{-1/2}X^\top\varepsilon\|_2.
$$
Suppose $\varepsilon$ has independent, mean-zero components that are
$\sigma_s^2$-sub-Gaussian.  Then $X^\top\varepsilon$ is a sub-Gaussian vector
with proxy covariance $\sigma_s^2 X^\top X$.
A standard concentration inequality for quadratic forms of sub-Gaussian vectors
(see, e.g., \citet{vershynin2025high}) implies that, for any $\delta_s\in(0,1)$,
with probability at least $1-\delta_s$,
$$
\|A_0^{-1/2}X^\top\varepsilon\|_2
\le
\sigma_s\left(
  \sqrt{\mathrm{tr}(X A_0^{-1} X^\top)}
  + \sqrt{2\|X A_0^{-1} X^\top\|_{\mathrm{op}} \log(1/\delta_s)}
\right).
$$
The leading trace term matches the scale of the variance contribution in
\eqref{eq:B0_expectation_main}, while the second term inflates this by an
operator-norm factor to account for rare large deviations.  In regimes where
the synthetic design is well-conditioned in the $A_0^{-1}$-geometry,
$\|X A_0^{-1} X^\top\|_{\mathrm{op}}$ is not much larger than
$\mathrm{tr}(X A_0^{-1} X^\top)$, so the noise contribution is sharply
concentrated around its mean.

\section{Model Information}
\label{app:model info}
\subsection{Experimental Details}

In our experiments, we train LinUCB \citep{chu2011contextual} with a fixed exploration parameter $\alpha = 10$. Data collection was performed using the OpenAI API \citep{openai2024gpt4technicalreport} and Hugging Face \texttt{transformers} library \citep{wolf2020huggingfacestransformersstateoftheartnatural} for open-weights models. 

All runs were executed on a 20-core Intel\textsuperscript{\textregistered} Core i7-14700F (2.1 GHz), 32 GB DDR5 RAM, and an NVIDIA GeForce RTX 4070 Ti SUPER GPU with 16 GB of dedicated memory. The largest full sweep tested, comprising the \texttt{10k} baseline, \texttt{10k\_x1}...\texttt{10k\_x7} variants, and a cold-start baseline, each repeated for ten independent rounds, completed in under two wall-clock hours.

\subsection{Model Specifications}
Table~\ref{tab:model_checkpoints} details the specific model revisions used. For open-weights models, we pinpoint the exact snapshot using the Hugging Face commit hash (first 7 characters). The earlier snapshot results in Table~\ref{tab:snapshot_gpt35_flip_old} correspond to the Jan--Feb 2025 period. 

\begin{table}[h]
    \centering
    \small
    \caption{Model specifications. OpenAI models are listed by access window; open-weights models include their Git revision ID.}
    \label{tab:model_checkpoints}
    \begin{tabular}{l|l|l|l}
        \toprule
        \textbf{Model} & \textbf{Variant} & \textbf{Checkpoint Path} & \textbf{Revision (Hash)} \\
        \midrule
        \textbf{Llama 3.1} & 8B Instruct & \texttt{meta-llama/llama-3.1-8B-Instruct} & \texttt{0e9e39f} \\
        \textbf{Qwen 3} & 8B Instruct & \texttt{Qwen/Qwen3-8B} & \texttt{b968826} \\
        \textbf{GPT-3.5} & Turbo & \textit{Proprietary API} & Sept--Oct 2025 \\
        \textbf{GPT-4o} & Omni & \textit{Proprietary API} & Sept--Oct 2025 \\
        \bottomrule
    \end{tabular}
\end{table}

\subsection{Inference Hyperparameters}
\label{app:inference_params}

To ensure fair comparison across model families, we aligned inference parameters as closely as possible. Table~\ref{tab:inference_settings} details the generation configuration. For OpenAI models, we utilized the default system settings with a fixed temperature. For open-weights models (Llama 3.1, Qwen 3), we utilized the \texttt{transformers} library with explicit generation limits to prevent infinite loops in chain-of-thought sequences.

\begin{table}[h]
    \centering
    \small
    \caption{Inference parameters.}
    \label{tab:inference_settings}
    \begin{tabular}{l|cc}
        \toprule
        \textbf{Parameter} & \textbf{OpenAI Models} & \textbf{Open-Weights Models} \\
                           & \textit{(GPT-3.5, GPT-4o)} & \textit{(Llama 3.1, Qwen 3)} \\
        \midrule
        Temperature        & 0.5                   & 0.5 \\
        Max Output Tokens  & Model Maximum         & 4,000 \\
        Top\_p             & 1.0                   & 1.0  \\
        Frequency Penalty  & 0.0                   & 0.0  \\
        Presence Penalty   & 0.0                   & 0.0  \\
        Stop Sequences     & None                  & None \\
        \bottomrule
    \end{tabular}
\end{table}

\subsection{Snapshot Sensitivity (GPT3.5 Turbo)}\
\label{app: snapshot sensitivity}
% change the thing
To assess the temporal stability of LLM-generated priors, we consider data generated with GPT3.5 Turbo using an earlier model snapshot (Jan--Feb 2025). All other components remain fixed (settings listed in Table~\ref{tab:inference_settings}). Table~\ref{tab:snapshot_gpt35_flip_old} reports the resulting $\delta$ regret under preference-flipping, in the same format as the main cross-domain summary tables, specifically compared to the results in Table~\ref{tab:full_results_all_models}. 
Comparing the two snapshots reveals significant performance degradation over time, suggesting a form of ``alignment drift.'' While the earlier snapshot achieved robust gains on the Immigration dataset ($4.03\%$ reduction at $N=10k$), the newer version reported in the main text (Table~\ref{tab:full_results_all_models}) failed to improve over cold-start ($0.04\%$ reduction). Similarly, on the Travel dataset, the warm-start benefit dropped from $4.11\%$ with the older model to $2.63\%$ with the newer version. On the COVID-19 dataset, while asymptotic performance ($N=10k$) was similar, the earlier snapshot provided a much stronger initialization at lower data regimes ($N=1k$), yielding a $17.57\%$ reduction compared to just $7.20\%$ for the newer model. These results indicate that effective alignment is not a static property of a model family (e.g., ``GPT-3.5'') but is sensitive to specific version updates and other adjustments.

\begin{table}
\centering
\small
\caption{GPT-3.5 Turbo results from older snapshot under preference flipping; percentage reduction in cumulative regret (\mbox{\%$\Delta$ Regret}) versus a cold-start LinUCB baseline. Mean over $G=10$ seeds \(\pm\) 95\,\% CI.}
\label{tab:snapshot_gpt35_flip_old}
\begin{tabular}{llrrr}
\toprule
\textbf{Dataset} & \textbf{Noise (\%)} & \textbf{N = 1k} & \textbf{N = 3k} & \textbf{N = 10k} \\
\midrule
COVID-19     & 0  & 17.57 ± 4.29 & 11.23 ± 2.61 & 9.45 ± 1.17 \\
COVID-19     & 30 & 11.27 ± 3.78 & 5.91 ± 4.48  & 7.44 ± 1.34 \\
COVID-19     & 50 & 7.74 ± 2.65  & 1.81 ± 4.20  & -8.07 ± 3.45 \\
Immigration  & 0  & 13.03 ± 2.37 & 6.22 ± 1.41  & 4.03 ± 1.22 \\
Immigration  & 30 & 6.49 ± 4.78  & 1.32 ± 2.83  & 0.56 ± 1.25 \\
Immigration  & 50 & -2.05 ± 4.16 & -15.15 ± 4.12& -17.33 ± 3.50 \\
Travel       & 0  & 4.46 ± 4.62  & 1.45 ± 2.67  & 4.11 ± 1.01 \\
Travel       & 30 & 1.39 ± 5.57  & 0.84 ± 0.84  & 0.18 ± 1.08 \\
Travel       & 50 & 0.42 ± 2.62  & 0.02 ± 3.78  & -0.51 ± 1.50 \\
\bottomrule
\end{tabular}
\end{table}

\subsection{Prompts}
The next subsections list the exact prompts used to generate synthetic conjoint responses. Placeholders such as \verb+[User]+ and \verb+[Vaccine A]+ are replaced at runtime.

\subsubsection{COVID-19 Vaccine}
Following \citet{alamdari2024jumpstartingbanditsllmgenerated}, we reuse their
prompt verbatim:
\begin{quote}\small
\textit{Consider you are in the middle of the COVID pandemic, where vaccines
are just being produced. Pretend to be the following user: [User]. Now you
are given two vaccine choices for COVID. The description of each vaccine is
as follows: [Vaccine A] Now the next one: [Vaccine B]. Which one do you
take? A or B? Let's think step by step. Print the final answer as
[Final Answer] at the end as well.}
\end{quote}

\subsubsection{Immigration}
\begin{quote}\small
\textit{Pretend to be the following user: [User]. You are now evaluating two
immigrants applying for admission to the United States. The description of
each immigrant is as follows: [Immigrant A] Now the next one: [Immigrant B].
Which immigrant do you admit? A or B? Let's think step by step. Print the
final answer as [Final Answer] at the end.}
\end{quote}

\subsubsection{Travel}
\begin{quote}\small
\textit{Consider you are planning a U.S. vacation and some states have
recently passed policies that weaken democratic principles. Pretend to be
the following user: [User]. Now you are given two locations for vacationing.
The description of each location is as follows: [Location A], now the next
one: [Location B]. Which location do you visit? A or B? Let's think step by
step. Print the final answer as [Final Answer].}
\end{quote} %first draft done. 
\end{document}